%% file: root.tex
\documentclass[letterpaper, 10 pt, conference]{ieeeconf}  

\IEEEoverridecommandlockouts                              

\usepackage{math}
\usepackage{amsmath}
\usepackage{flushend}

\title{\LARGE \bf GND: Global Navigation Dataset with Multi-Modal Perception and Multi-Category Traversability in Outdoor Campus Environments}

\author{ Jing Liang*$^{1}$, Dibyendu Das*$^{2}$, Daeun Song*$^{2}$, Md Nahid Hasan Shuvo$^{2}$, Mohammad Durrani$^{1}$, \\
Karthik Taranath$^{1}$, 
Ivan Penskiy$^{1}$, Dinesh Manocha$^{1}$, Xuesu Xiao$^{2}$
\thanks{$^{1}$University of Maryland, College Park. $^{2}$George Mason University.}
\thanks{*Equally contributing authors.}
}

\newcommand{\name}[1]{\parbox{1.7cm}{\centering #1}}
\newcommand{\sensors}[1]{\parbox{5.7cm}{\centering #1}}
\newcommand{\purpose}[1]{\parbox{2.6cm}{\centering #1}}

\begin{document}

\maketitle
\thispagestyle{empty}
\pagestyle{empty}


\input{v1}

\newpage
\bibliographystyle{IEEEtran}
\bibliography{ref}


\end{document}

%% file: v1.tex
\begin{abstract}
Navigating large-scale outdoor environments requires complex reasoning in terms of geometric structures, environmental semantics, and terrain characteristics, which are typically captured by onboard sensors such as LiDAR and cameras. While current mobile robots can navigate such environments using pre-defined, high-precision maps based on hand-crafted rules catered for the specific environment, they lack commonsense reasoning capabilities, especially the traversability analysis, that most humans possess when navigating unknown outdoor spaces. To address this gap, we introduce the Global Navigation Dataset (GND), a large-scale dataset that integrates multi-modal sensory data, including 3D LiDAR point clouds and RGB and 360° images, as well as multi-category traversability maps (pedestrian walkways, vehicle roadways, stairs, off-road terrain, and obstacles) from ten university campuses.  These environments encompass a variety of parks, urban settings, elevation changes, and campus layouts of different scales. The dataset covers approximately 2.7km$^2$ and includes at least 350 buildings in total. We also present a set of novel applications of GND to showcase its utility to enable global robot navigation, such as map-based global navigation, mapless navigation, and global place recognition. GND's website can be found at \url{https://cs.gmu.edu/~xiao/Research/GND/}.



\end{abstract}

\section{Introduction}
\label{sec:introduction}
Global navigation plays a critical role in enabling robots to traverse large-scale outdoor environments~\cite{congram2021relatively, liu2021team, ort2018autonomous, liang2024dtg, liang2024mtg}. It is widely used in real-world tasks like last-mile delivery~\cite{hoffmann2018regulatory, chen2021adoption}, remote exploration~\cite{cao2023representation, cao2021tare}, autonomous driving~\cite{ort2018autonomous, yurtsever2020survey}, etc. Unlike navigation in structured indoor spaces~\cite{xiao2023autonomous, park2023learning, xiao2022autonomous, stein2018learning}, global navigation needs to reason about a variety of environmental factors in complex outdoor scenarios~\cite{liang2022adaptiveon, weerakoon2023graspe, liang2024dtg, shah2022viking}, including recognizing terrain characteristics for traversability analysis~\cite{datar2023learning, datar2024terrain, datar2024toward, xiao2021learning, karnan2022vi} and inferring navigational cues to determine the shortest path in large-scale open environments~\cite{stein2018learning, liu2021team}.

One challenge of global navigation is the need of navigational reasoning at a very large scale, e.g., trajectories corresponding to hundreds or thousands of meters~\cite{congram2021relatively, liu2021team, ort2018autonomous, liang2024dtg, liang2024mtg}. In practice, navigating from the south side of a university campus to a cafeteria on the north side without a prior map may require following major pedestrian walkways at the beginning, while taking shortcuts close to buildings later. Such a reasoning process requires not only geometric, but also semantic understanding of the large-scale outdoor scenes, which needs to be captured by various sensors like 3D LiDARs and RGB cameras. 

\begin{figure}
    \centering
    \includegraphics[width=\linewidth]{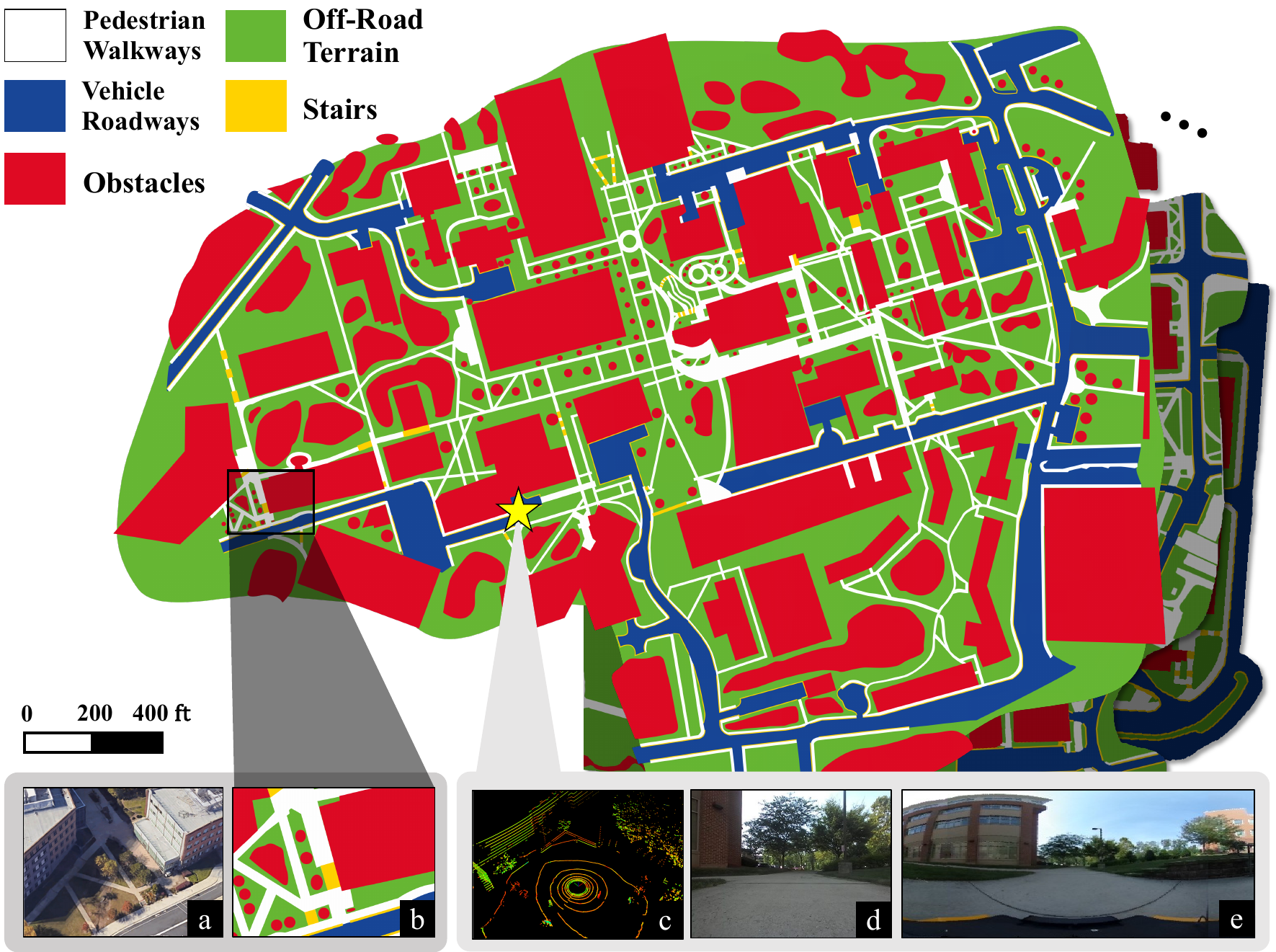}
    \caption{\textbf{Traversability maps of the Global Navigation Dataset (GND)}: The legend shows five different categories of terrain traversability with different colors. Inset (a) shows the 3D satellite image of inset (b), an enlarged traversability map. GND contains multi-modal sensory data including 3D LiDAR point clouds (c), RGB images (d), and 360\textdegree~images (e).}
    \label{fig:enter-label}
    \vspace{-2.0em}
\end{figure} 

Another challenge is making navigation decisions by reasoning beyond simple obstacles and free spaces, a common delineation of indoor workspaces, while considering different robot morphologies. For example, pedestrian walkways, vehicle roadways, and some off-road terrain may all appear as free spaces, but they may correspond to different categories of traversability in different scenarios~\cite{sikand2022visual}, e.g., stairs should be treated as obstacles for wheeled robots, but they can be regarded as free spaces for legged robots~\cite{liu2021team}. 

Although robotics practitioners have tackled these challenges with high-precision prior maps and hand-crafted navigation rules for specific environments, recent research has focused on leveraging machine learning to equip robots with generalizable, human-like reasoning capabilities during outdoor global navigation. Imitation and reinforcement learning techniques~\cite{liang2022adaptiveon, weerakoon2023graspe} have been used for semantic understanding~\cite{hosseinpoor2021traversability,oh2022travel, dabbiru2021traversability, cao2023representation, roddick2020predicting, ma2019accurate}, traversability analysis~\cite{hosseinpoor2021traversability,oh2022travel, dabbiru2021traversability, cao2023representation,roddick2020predicting, ma2019accurate}, topological modeling~\cite{shah2022viking, shah2023vint}, trajectory generation~\cite{liang2024mtg, liang2024dtg}, heuristic estimation~\cite{stein2018learning, stein2021generating}, parameter tuning~\cite{xiao2022appl, xiao2020appld, wang2021appli, wang2021apple, xu2021applr}, and policy learning~\cite{pfeiffer2017perception, francis2020long}. One common requirement of all these data-driven approaches is high-quality ground truth or trial-and-error data for training.

\input{datasets}
\noindent {\bf Main Results:} Motivated by such difficulties of global navigation and research needs of training data, we present a novel, large-scale Global Navigation Dataset (GND), which includes multi-modal perception and multi-category traversability in outdoor campus environments (Fig.~\ref{fig:enter-label}). 
GND comprises almost 11 hours of navigation data captured using two Clearpath Jackal robots, accompanied by 3D LiDAR point clouds, RGB and 360\textdegree~camera images, inertia Measurement Unit (IMU) information, GPS data, as well as robot odometry and actions, collected across \textbf{10} university campuses in city and village areas, including a variety of park areas, vegetation types, elevation changes, diverse campus layouts and objects in the campuses. In total, We covered around \textbf{2.7km$^2$} with at least \textbf{350} buildings in the datasets over 11 hours of recorded data (in rosbags). 
All raw perception data are post-processed into ten large-scale global campus maps labeled with five categories of traversability (pedestrian walkways, vehicle roadways, stairs, off-road terrain, and obstacles) and associated with multi-modal perception (e.g., first-person and 360\textdegree~view) on the robot trajectories. Some of our main contributions include: 
\begin{itemize}
  \item 
  The first large-scale, long-range, across-campus global navigation dataset with multi-modal perception data and multi-category traversability maps;   
  \item A standardized and streamlined data collection and post-processing pipeline designed to encourage broader contributions from all users to the dataset; and
  \item Novel dataset applications showing GND's utility in enabling outdoor global navigation tasks with different types of robots (wheeled and legged robots), as shown in Section~\ref{sec:usecases}, including global map-based navigation (path planing), mapless navigation (trajectory and motion generation), and global place recognition. 
\end{itemize}


\section{Related Work}
\label{sec:related_works}

In this section, we review related literature on global robot navigation and state-of-the-art robot navigation datasets.

\subsection{Global Robot Navigation}

Navigating robots in large, outdoor environments presents multiple challenges, including the need to assess terrain traversability and perform large-scale navigational reasoning. Global robot navigation can be divided into two main approaches: map-based and mapless. Map-based approaches rely on a comprehensive cost map for path planning~\cite{fusic2021optimal, li2021openstreetmap} and precise robot localization~\cite{akai2017robust, lowry2015visual} to ensure accurate path execution. However, these methods can be computationally expensive and require significant overhead to maintain up-to-date maps. To address these limitations, ViNT~\cite{shah2023vint} and NoMaD~\cite{sridhar2023nomad} proposed generating topological maps and using vision-based images as subgoals for navigation, though they still require initial runs to collect subgoal images. On the other hand, mapless navigation techniques eliminate the need for maps entirely. AdaptiveON~\cite{liang2022adaptiveon} focused on generating actions in a mapless fashion but was limited to local planning without addressing long-distance navigation. More recent advancements, such as MTG~\cite{liang2024mtg}, enable long-distance navigation, while DTG~\cite{liang2024dtg} further optimizes traversable trajectories in large-scale outdoor settings. 
Both map-based and mapless approaches require extensive datasets with highly accurate traversability maps and multi-modal sensory data for effective reasoning and training.


\subsection{Datasets for Robot Navigation}

Over the past decade, large-scale navigational datasets have proven invaluable across various research domains, including social robot navigation~\cite{hirose2023sacson}, human trajectory prediction~\cite{thor}, autonomous driving~\cite{ort2018autonomous, yurtsever2020survey}, vision-based navigation~\cite{shah2023gnm}, and global navigation~\cite{liang2024dtg}.

\input{table2}

KITTI~\cite{kitti} was one of the first large-scale datasets that emphasized the importance of well-organized, real-world data for advancing machine learning and computer vision research. It not only significantly impacted autonomous driving but also influenced the broader field of computer vision. Although many follow-up datasets have been introduced~\cite{nuscenes, waymo, cordts2016cityscapes}, early efforts were predominantly focused on autonomous vehicle applications, particularly in perception. 
Similarly, in robotics, multi-sensory datasets have been released. However, current datasets focused on perception or mapping tasks~\cite{flobot, martin2021jrdb, ncarlevaris-2015a}, considered specific local navigation tasks~\cite{scand, musohu, sit}, or were limited in scale and sensor modalities~\cite{shah2021rapid}. As a result, there remains a gap in the datasets among adequately covering large-scale environments and traversability analysis across different sensor and robot modalities. GND addresses this gap by offering rich multi-modal robot sensory data collected in outdoor campus environments, complemented by human-labeled multi-category traversability maps. Table~\ref{table:comp} compares our work with relevant state-of-the-art datasets.

While previous datasets have proven to be valuable for studying perception and various navigation challenges, they fall short in providing demonstrations on global navigation, particularly regarding traversability information. While SiT~\cite{sit} offered 12-layered semantic maps, its primary focus was on providing benchmarks and analysis for pedestrian detection and tracking, falling short in demonstrating practical applications that leverage the semantic maps.  We aim to showcase applications of GND by utilizing various data and methods to demonstrate their practical applications.


\begin{figure}[tb]
\centering

\includegraphics[width=.8\linewidth]{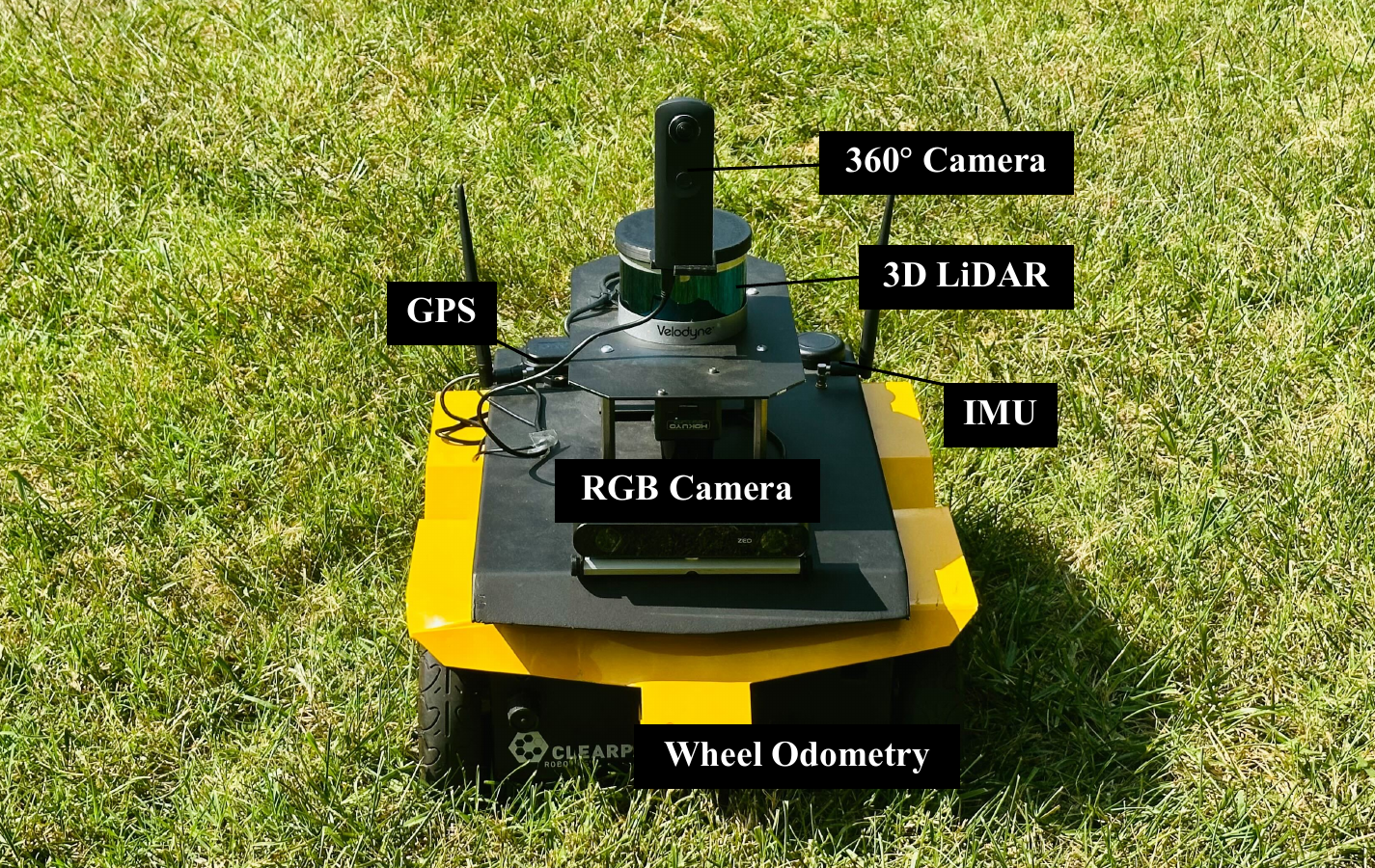}
\caption{\textbf{Robot Setup}: We use a Clearpath Jackal for dataset collection, which is equipped with various sensors, including 3D LiDAR, RGB camera, 360\textdegree~camera, IMU, and GPS. It is capable of traversing diverse terrains, including pedestrian roads, roadways, off-road areas, ramps, and woods.
}
\label{fig:robot} 
\vspace{-1.5em}
\end{figure}

\section{Dataset}
\label{sec:dataset}

In this section, we first describe the data collection procedure. We then describe the details of our dataset, particularly on the traversability map. 

\subsection{Data Collection}

We manually operate the robot to navigate various campus environments for data collection. We guide the robot considering the traversability of the road. The robot primarily navigates pedestrian walkways; However, when necessary, it also traverses vehicle roadways, such as when crossing streets or accessing specific areas. As shown in Fig.~\ref{fig:robot}, the robot is equipped with the following sensors:
\begin{itemize}
  \item \textbf{3D LiDAR}: Velodyne VLP-16 with 16 channels or Ouster OS1-32 with 32 channels, both covering a 360-degree field of view and operating at 10 Hz;
  \item \textbf{RGB Camera}: ZED2 with image resolution of $1080p$ facing front and operating at 15 Hz; 
  \item \textbf{360\textdegree~Camera}: RICOH Theta V operating at 15 Hz;
  \item \textbf{IMU}: 6D 3DM-GX5-10 operating at 355 Hz; and
  \item \textbf{GPS}: u-blox F9P operating at 20 Hz.
\end{itemize}


Our robot operates on Ubuntu 20.04 and Robot Operating System (ROS) Noetic. The data captured by the sensors are recorded in the rosbag file format. We also provide both intrinsic and extrinsic calibration parameters for the LiDARs and the cameras. We gathered datasets from 10 university campuses with approximately 2.7 km$^2$ of campus area, including at least 350 buildings, with over 11 hours of recorded rosbag data. These campus datasets encompass a variety of environments, such as parks, different types of vegetation, elevation changes, diverse campus layouts, and objects. As shown in Table~\ref{table:gnd}, we list five example campus datasets with their details. The campus datasets cover a various size of campus areas ranging from 0.15 km$^2$ to 0.84 km$^2$, with 32 to 60 buildings. The table also underscores the variety of traversability category ratios in different campuses. For examples, the pedestrian walkways range from 7.16$\%$ to 17.31$\%$ over the covered campus area. More campus datasets are in the website in the Abstract.

\begin{figure*}[ht]
\centering
\includegraphics[width=\textwidth]{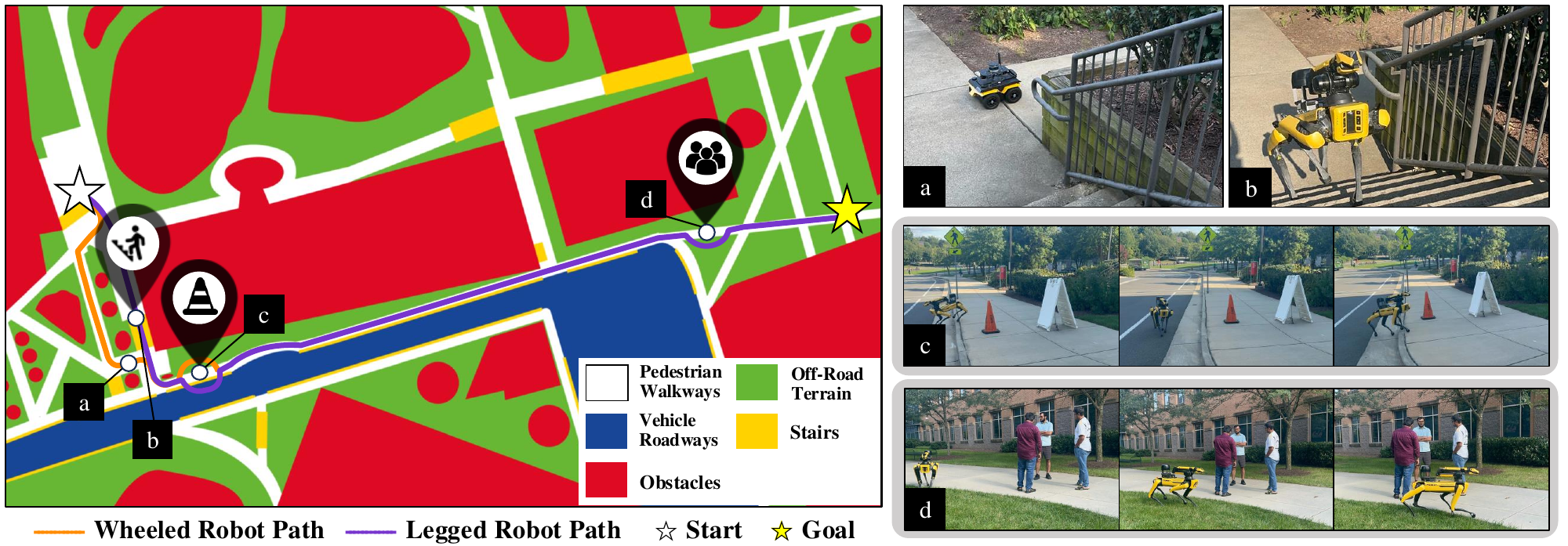}
\caption{\textbf{Map-based Global Navigation using Multi-Category Traversability Map in GND}: The white and yellow stars indicate the start and goal positions, respectively. The orange line shows the path of the wheeled robot, and the purple line shows the path of the legged robot. (a) and (b) shows Scenario 1, where there are stairs. (c) shows Scenario 2, where the robot is going to the right bottom of the figure. When the road is blocked and the legged robot can only go through the roadway due to narrow passages in off-road terrain. (d) shows Scenario 3, when the road is blocked and the robots have to go around the path using the off-road terrain.}
\label{fig:inno1}
    \vspace{-1.7em}
\end{figure*}

\begin{figure}[tb]
\centering
\subfigure[Point Cloud Map]{
\includegraphics[width=0.48\linewidth]{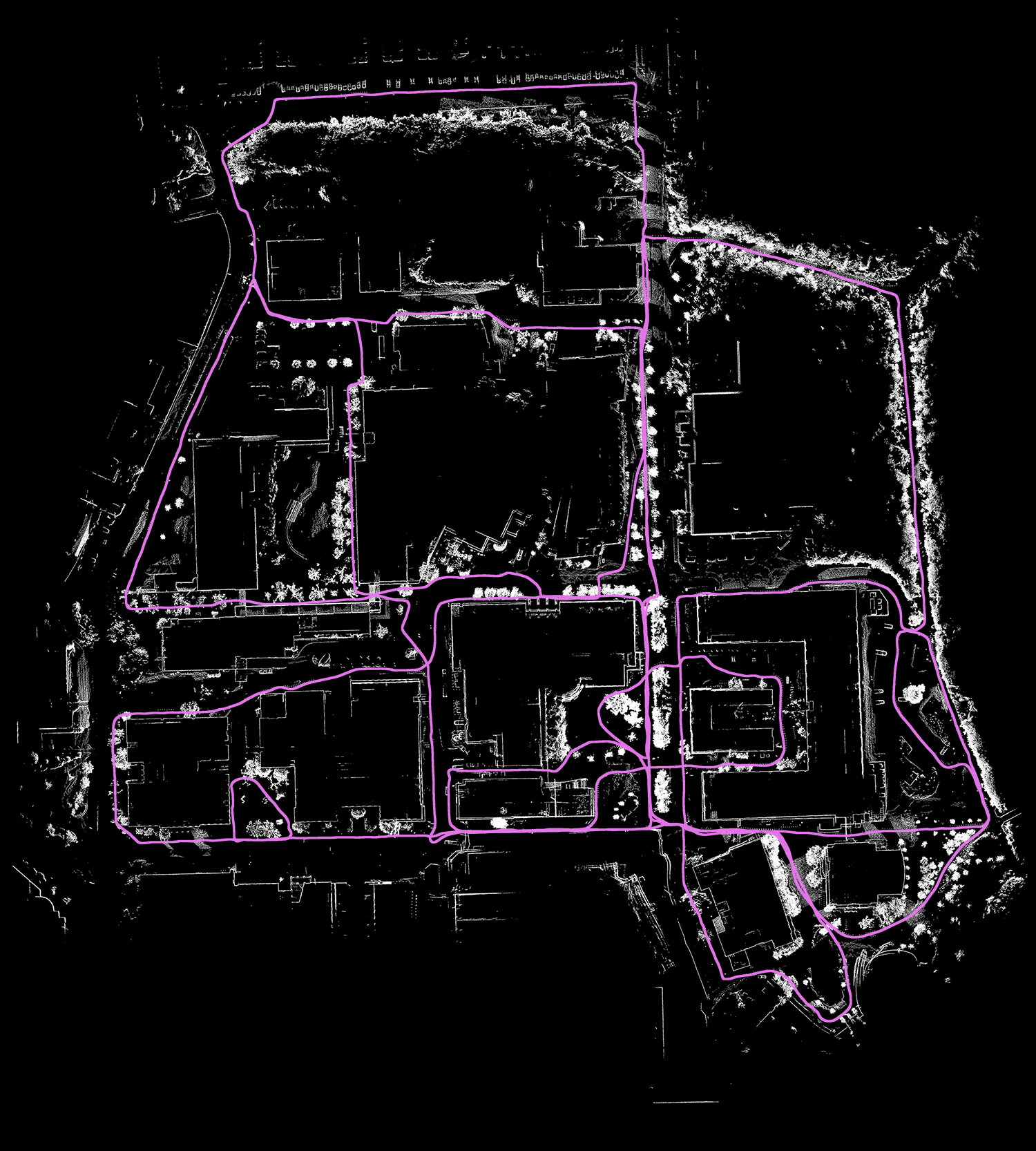}
}\hspace{-0.8em}
\subfigure[Multi-category Traversability Map]{
\includegraphics[width=0.48\linewidth]{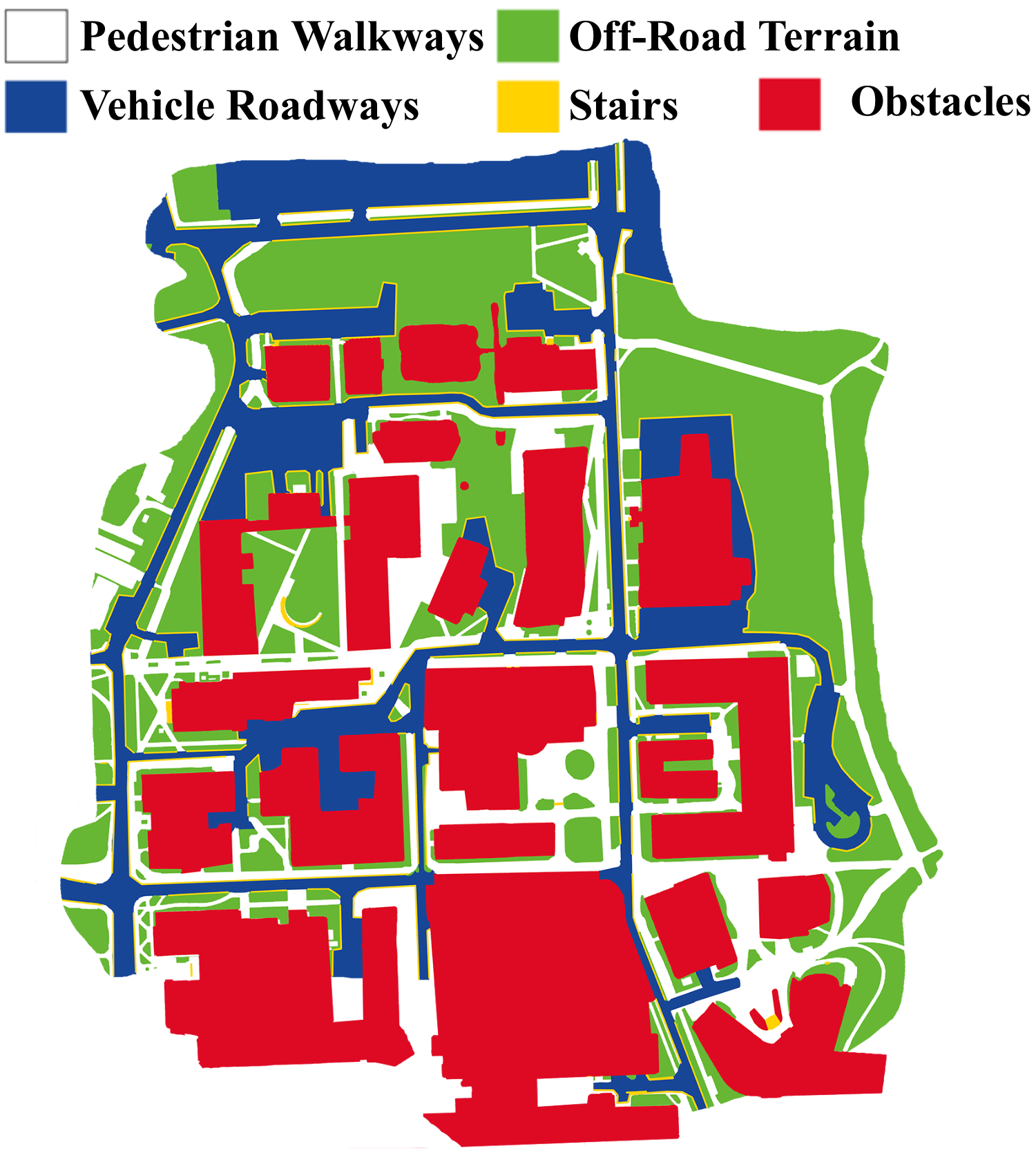}
}
\vspace{-0.7em}
\caption{\textbf{Partial Map of the University of Maryland}: Inset (a) presents the processed point cloud map, featuring the robot's trajectory for data collection marked by a pink line; (b) illustrates the multi-category traversability map corresponding to the dataset shown in (a).
}\label{fig:map}
\vspace{-1.5em}
\end{figure}

\subsection{Standardized Data Processing}

To encourage broader contributions from dataset users, we standardize our data processing workflow. First, the raw rosbag data is processed using LIO-SAM~\cite{shan2020lio} or FAST-LIO~\cite{xu2021fast} to generate both the trajectories and 3D local maps. Next, we process the point cloud maps by removing the ground, which enhances the visibility of significant features, such as buildings, in a top-down view. This ground removal also improves localization performance. Then, these local maps are registered using TEASER++~\cite{yang2020teaser} to create a global map, where all trajectories and maps are transformed into the global map's coordinate system. For each campus, we generate a single global map within the dataset. This global map is a comprehensive 3D representation of the entire campus, designed for use in map-based navigation approaches and as ground truth for mapless ones. 

To create 2D traversability maps, we project the 3D point cloud global map onto the 2D plane along the Z-axis. A standard annotation method is then applied to label five distinct traversability types, each represented by different colors. The standardized data processing pipeline is also published with the paper.

\subsection{Multi-Category Traversability Map}

The multi-category traversability map consists of five traversability types. White indicates areas traversable by most robots, such as sidewalks, concrete surfaces, and brick roads. Red represents areas that are non-traversable for all robots, including buildings, rivers, construction sites, and poles. Between these extremes, we define three additional categories: green marks off-road or vegetated areas, yellow represents stairs or curbs, and blue highlights roadways and parking lots. Different types of robots can navigate through different traversable areas. For example, legged robots can handle stairs and curbs, while wheeled robots cannot. However, fast-moving wheeled robots are capable of traveling on roadways, whereas slow legged robots are not safe in high-traffic environments.

Figure~\ref{fig:map} shows an example of the multi-category traversability map, alongside the point cloud map. It demonstrates that GND provides not only geometric but also semantic information about the environment, closely aligned with real-world conditions.


\input{extended_mtg}

\section{Applications}
\label{sec:usecases}

In this section, we present three applications for the GND dataset, emphasizing its unique characteristics: \emph{globalness} and \emph{traversability}, which do not present in existing navigation datasets. This dataset is collected mostly by Jackal robot, but it can be used for navigation tasks with different types of robots, such as legged robots and wheeled robots. We implement map-based global navigation, mapless navigation, and global place recognition.

\subsection{Map-based Global Navigation}

The primary objective of the GND dataset is to provide precise map data for global robot navigation. To demonstrate its utility, we conduct an experiment comparing the navigation of two robots with different modalities and traversabilities, wheeled and legged. Using the map, path planning methods such as A* or RRT* can generate a path based on the GPS coordinates of the start and goal positions. As the robot moves, motion planning methods like the classical approaches~\cite{fox1997dynamic,liang2021vo}, like Dynamic Window Approach~\cite{fox1997dynamic}, or learning-based approaches~\cite{liang2022adaptiveon, liang2021crowd, das2024motion, xiao2021toward, xiao2021agile, wang2021agile} can be employed to observe the real-time environment changes and guide the robot's actions. Both robots will initially navigate along the sidewalk, but if the path becomes non-traversable for a particular robot type, the motion planner will select alternative traversable areas, adjusting the robot's course to reach the next waypoint along the trajectory.

As shown in Figure~\ref{fig:inno1}, the traversability map illustrates various scenarios during the robots' navigation from the white star to the yellow star. The purple trajectory represents the path of the legged robot, while the orange trajectory shows the wheeled robot's path. When the path encounters stairs, the wheeled robot deviates to a nearby ramp before returning to the next waypoint, as depicted in the RGB image $(a)$ on the right side of the figure. Meanwhile, the legged robot continues on its original path, walking directly up the stairs. For other obstacles, such as construction cones and groups of people blocking the sidewalk, as shown on the right side of Figure~\ref{fig:inno1}, the legged robot steps down the curb or navigates through off-road terrain to avoid the blockage.



\subsection{Mapless Navigation with Traversability Analysis}

\begin{figure}[tb]
    \centering
    \includegraphics[width=\linewidth]{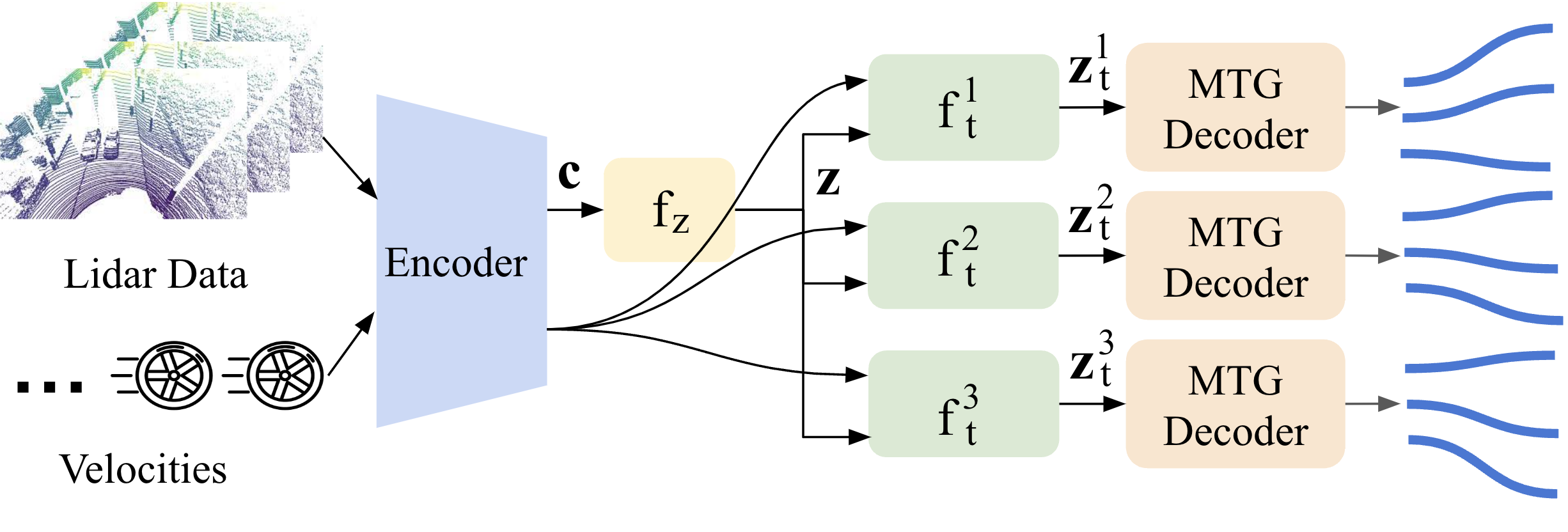}
    \caption{\textbf{T-MTG}: As shown in Equation~\ref{eq:ext_mtg}, for different traversabilities, we generate corresponding embedding vectors $\z_t^{k}$ from $\z$, under the condition of observation $\o$. The MTG Decoder is used to decode the embedding vectors into trajectories. Finally, the output of the model includes trajectories in all the traversability levels.} 
    \vspace{-1.5em}
    \label{fig:t-mtg}
\end{figure}

To assess the efficacy of various traversability types in the dataset for learning-based mapless navigation algorithms, we extend the MTG algorithm~\cite{liang2024mtg} with multiple traversability levels, referred to as T-MTG (Fig.~\ref{fig:t-mtg}). The problem formulation in MTG is given by Equation~\ref{eq:mtg}:
\begin{align}
    p(\tau|\c) &\approx \frac{1}{S}\sum_{s=1}^S p(\tau|\z^{(s)},\c), \;\;\;\z^{(s)} \sim p_\theta(\z|\o),
    \label{eq:mtg}
\end{align}
where $\tau$ represents the generated trajectories under the condition $\c=f_c(\o)$. $f_c(\cdot)$ is a sequence of linear layers, and $\o$ denotes the observation information. Here, $\z=f_z(\c)$ is the embedding vector of the encoded observation and $f_z(\cdot)$ represents a sequence of linear layers. During training, $\z = \set{\z^{(s)}}$ is sampled from the distribution $p_\theta(\z|\o)$, where $p_\theta(\cdot)$ is the distribution of $\z$ and $\theta$ represents the parameters of the encoder model. $S$ indicates the number of waypoints.

\begin{figure*}[t]
    \centering
    \includegraphics[width=0.95\textwidth]{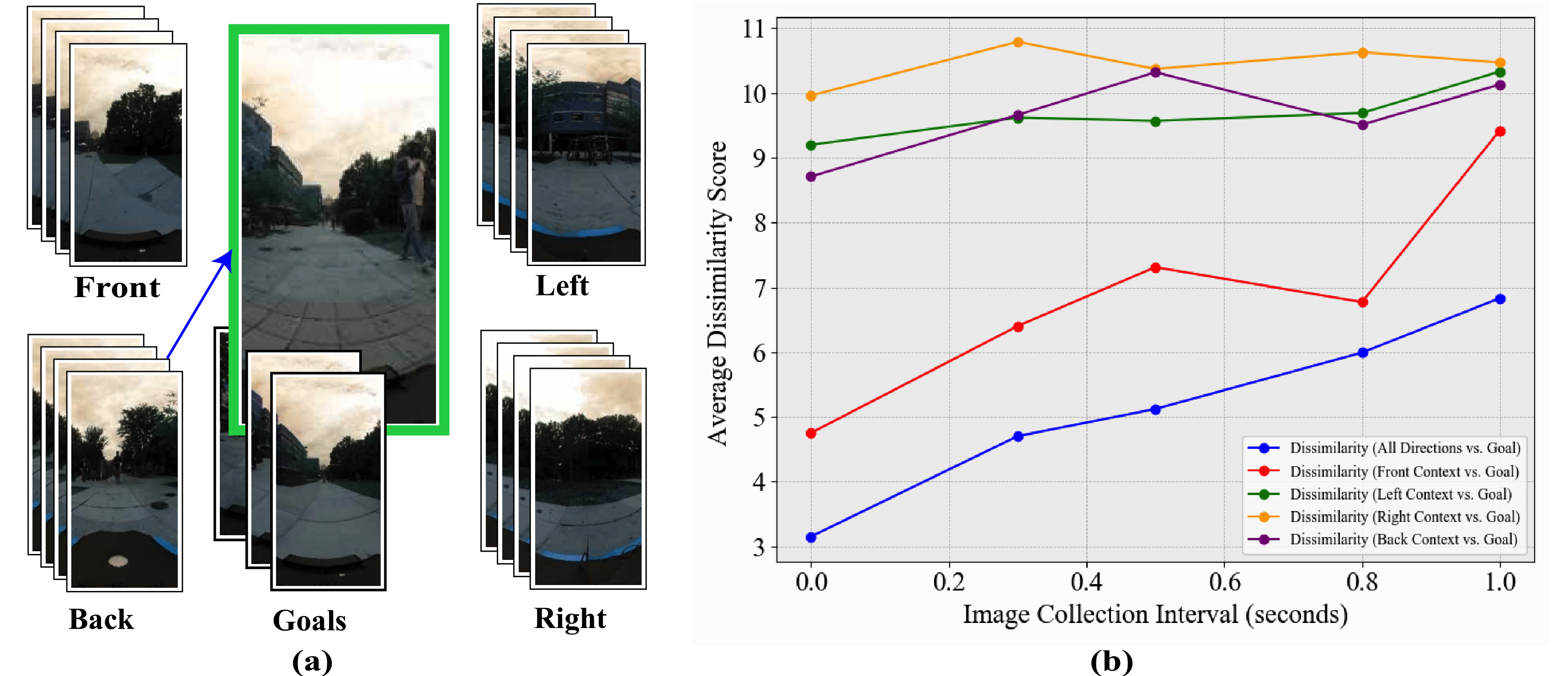}
    \caption{\textbf{Vision-based Place Recognition using 360\textdegree~images in GND}: Inset (a) shows the application of 360\textdegree~context images for goal detection. In this particular example, when all four directional context images are available, the back context image is selected as the closest to the goal image (green-bordered), highlighting the advantage of 360\textdegree~images in goal detection. (b) represents how the average dissimilarity between all directional context images and the goal (blue line) decreases compared to using only one directional context image.}
    \vspace{-1.5em}
    \label{fig:Inno3}
\end{figure*} 

For different traversability levels, we aim for the observation embeddings to capture the current traversabilities. To achieve this, a model is employed to reprocess the traversability: $\z_t = f_t^k(\z_t|\z, \o)$, where $\o$ is reused to provide residual information, enhancing the calculation of $\z_t$.  $k$ represents the traversability level. Thus, the MTG model in Equation \ref{eq:mtg} is extended to the T-MTG model in Equation~\ref{eq:ext_mtg}:
\begin{align} 
p(\tau_k|\c) &\approx \frac{1}{S}\sum_{s=1}^S p(\tau_k|\z_t^{k,(s)},\c), \nonumber \\
\z_t^{k} &\sim p_\theta(\z_t^k | \f_z(\z|\o), \o),  
\label{eq:ext_mtg}
\end{align}
where $\f_z(\cdot)$ represents the encoder and $\z_t^{k,(s)}$ is the embedding of the waypoint $s$ in $\z_t^k$.


As shown in Fig.~\ref{tb:extended_mtg}, we implement three traversability levels: Basic traversability includes only pedestrian walkways, where robots can move in various speed on in the areas; agile traversability level is designed primarily for fast-moving wheeled robots and includes both pedestrian walkways and vehicle roadways, where the robot is required to move fast to keep up with traffic; and legged traversability level is suited for legged robots, allowing traversal on pedestrian walkways and off-road terrain, though it is not safe for use on vehicle roadways. Our approach, T-MTG, generates trajectories to cover the 200\textdegree~field of view (FOV) in front of the robot. In Fig.~\ref{tb:extended_mtg}, the RGB images display the generated trajectories from the front camera's perspective (70\textdegree~FOV), and the traversability map highlights each waypoint of the trajectories in their respective traversability regions. For each traversability level, our approach successfully generates trajectories that lie within the appropriate regions, effectively covering the areas in front of the robot.

\subsection{Vision-based Place Recognition}

We collect both RGB and 360\textdegree~camera images to offer vision-based global navigation in the GND dataset. To demonstrate the usability of 360\textdegree~image data, we conduct an experiment using the NoMaD~\cite{sridhar2023nomad} algorithm for goal detection, which compares the current observation with topological image nodes to recognize the best target to follow the recorded topological nodes. For goal-directed navigation, NoMaD~\cite{sridhar2023nomad} encodes images of the robot’s current RGB observations as vectors, and then uses the vectors to predict the temporal distance to the goal, by calculating the similarities with the vectors of the sub-goal images of the topological nodes. The sub-goal image with the highest similarity is then chosen as the closest node for goal-directed navigation. 

We use images in four views (front, left, right, and back) from the 360\textdegree~camera to implement the NoMaD~\cite{sridhar2023nomad} algorithm. Initially, we collect images with different time intervals to generate a sparse topological map. The topological map is constructed using only the front view of the 360\textdegree~images.  Then we use the four views from the 360\textdegree~images as the robot's current observations to generate four embedded vectors.
Among these vectors, we determine the direction of the context images that has the closest distance to the sub-goals and select the subgoal with the highest similarity as the closest node for further navigation, as shown in Fig.~\ref{fig:Inno3}(a).

We notice that the average similarity score of all four views from the real-time image observations and the goal (blue line) is higher than the comparisons with only of , as shown in Fig.~\ref{fig:Inno3}(b). Additionally, the temporal distance between observed images and sub-goals also increases with a sparser topological map. These findings highlight the advantages of utilizing all available directional information from 360\textdegree~images for more accurate and efficient goal-directed navigation.


\section{Conclusion, Limitations, and Future Work}

We introduce GND, a large-scale, long-range, global robot navigation dataset that includes multi-modal perception data and multi-category traversability maps. This dataset enables robots to undertake long-range navigation while accounting for geometric and semantic traversability. We are making all data available online, accompanied by a standardized post-processing pipeline, and we encourage contributions from the wider research community to enhance the dataset. The primary features of GND—\emph{globalness} and \emph{traversability}—are illustrated through three distinct experiments, highlighting its potential applications. Although GND offers valuable sensory data, we recognize that there is still work to be done in providing benchmark scenarios and evaluation metrics for global robot navigation, which remain underexplored in the existing literature.

\noindent\textbf{Acknowledgment:} This work was supported in part by ARO Grant W911NF2310352 and Army Cooperative Agreement W911NF2120076

%% file: datasets.tex
\begin{table*}[ht] 
\centering
\caption{\textbf{Comparison of State-of-the-Art Navigation Datasets}: Our GND dataset features the most comprehensive range of sensors and provides multi-category traversability maps for global navigation in various campuses.}
\setlength{\tabcolsep}{5pt}
\begin{tabular}{ccccccc} 
 \toprule 
\multirow{2}{*}{\textbf{Dataset}} & \textbf{Traversability}& \textbf{Distance} & \textbf{Duration} & \multirow{2}{*}{\textbf{Sensors}} & \multirow{2}{*}{\textbf{Platform}} & \multirow{2}{*}{\textbf{Purpose}} \\
 & \textbf{Labels} & \textbf{(km)} & \textbf{(min.)} &  &  & \\
 \midrule
 \name{SCAND\cite{scand}} & \ding{53} & 40 & 522 & \sensors{3D LiDAR, RGB Camera, RGB-D Camera, \\Wheel Odometry, Visual Odometry}   & Robot & \purpose{Social Navigation} \\
 \midrule
 \name{MuSoHu\cite{musohu}} &  \ding{53}  & 100 & 1200 & \sensors{3D LiDAR, RGB-D Camera, 360\textdegree~Camera, \\IMU, Microphone, Visual Odometry}   & Human & \purpose{Social Navigation} \\
 \midrule
 \name{SiT\cite{sit}} &\checkmark & N/A & 24.3 & \sensors{3D LiDAR, RGB Camera (covering 360\textdegree), \\IMU, GPS, Wheel Odometry} & Robot & \purpose{Social Navigation, Human Detection \\ \& Tracking} \\
 \midrule
 \name{JRDB\cite{martin2021jrdb}} & \ding{53}  & N/A & 64 & \sensors{3D LiDAR, 2D LiDAR, RGB Camera, RGB-D Camera, 360\textdegree~Camera, IMU, GPS, Microphone, Wheel Odometry}   & Robot & \purpose{Human Detection \\ \& Tracking} \\
 \midrule
 \name{NCLT\cite{carlevaris2016university}}&\ding{53} & 147.4  & 2094  & \sensors{3D LiDAR, RGB Camera, Fisheye Camera, \\IMU, GPS, Wheel Odometry}   & Robot & \purpose{Global Navigation} \\
 \midrule
 \name{GND (Ours)}& \checkmark  & 53  & 668 & \sensors{3D LiDAR, RGB Camera, 360\textdegree~Camera, \\IMU, GPS, Wheel Odometry}   & Robot & \purpose{Global Navigation} \\
 \bottomrule
\end{tabular}
\label{table:comp}
    \vspace{-2em}
\end{table*}

%% file: table2.tex
\begin{table*}[t] 
\centering
\caption{\textbf{Five Example Campuses in GND}: The table outlines details of five example campuses, including the University of Maryland (UMD), George Mason University (GMU), Catholic University of America (CUA), Georgetown University, and George Washington University (GWU). We list the covered areas, number of buildings, trajectory length, number of RGB and 360\textdegree~images, number of LiDAR point clouds, and ratio of different traversability categories in the campus map. P, O, V, and S represent pedestrian walkways, off-road terrain, vehicle roadways, and stairs, respectively. For example, 0.84 km$^2$  is covered on the UMD  campus with 60 buildings, and the ratio of pedestrian walkways in the campus is $10.66\%$ of the UMD campus map. More campus datasets are on GND's website.}
\begin{tabular}{ccccccccccc} 
 \toprule 
\multirow{2}{*}{\textbf{Campuses}} & \textbf{Covered}      & \textbf{$\#$ of }& \textbf{Trajectory} & \textbf{$\#$ of RGB}    &\textbf{$\#$ of  360\textdegree} & \textbf{$\#$ of LiDAR} & \multicolumn{4}{c}{\textbf{Ratio of Traversability ($\%$)}} \\
& \textbf{Areas (km$^2$)} & \textbf{Buildings} & \textbf{Length (km)}& \textbf{Images} & \textbf{Images} & \textbf{Clouds} & P & O & V & S\\
 \midrule
 \name{UMD} & 0.84 & 60        & 23.26 & 214768 & N/A & 146703 & 10.66 & 16.29 & 25.84 & 1.67  \\
 \name{GMU} & 0.46 & 51        & 13.67 & 137948 & 137027 & 91500  & 17.31 & 25.04 & 17.11 & 0.41  \\
 \name{CUA} & 0.40 & 32        & 2.87 & 29921 & 30266 & 20025  & 7.86  & 42.29 & 18.78 & 1.81  \\
 \name{Georgetown} & 0.25 & 40 & 3.25 & 33244 & 33325 & 22050  & 7.16  & 21.42 & 13.96 & 1.51  \\
 \name{GWU} & 0.15 & 39        & 3.00 & 33156 & 32714 & 22190  & 8.95  & 14.04 & 28.09 & 1.99  \\
 \bottomrule
\end{tabular}
\label{table:gnd}
    \vspace{-1.5em}
\end{table*} 

%% file: extended_mtg.tex
\begingroup
\setlength{\tabcolsep}{2pt}

\begin{figure*}[!ht]
  \centering
  \begin{tabular}{cc|cc|cc}
    
   \multicolumn{2}{c|}{\textbf{Basic Traversability} } & \multicolumn{2}{c|}{\textbf{Agile Traversability}} & \multicolumn{2}{c}{\textbf{Legged Traversability}}\\ 
   \hline
    \includegraphics[height=.13\linewidth]{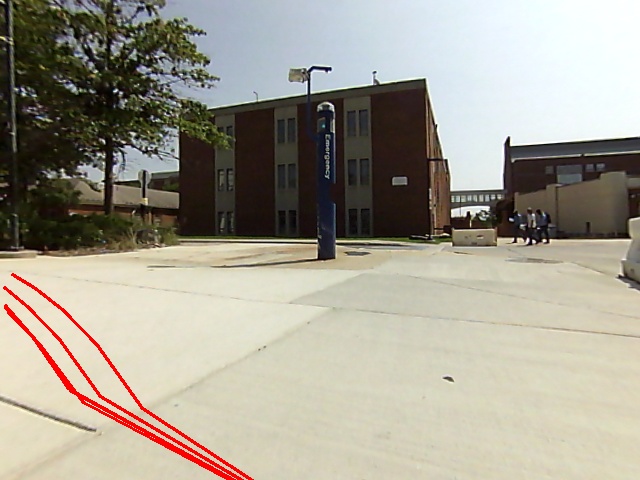}
    &
    \includegraphics[height=.13\linewidth]{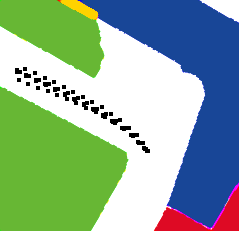} 
    &
    \includegraphics[height=.13\linewidth]{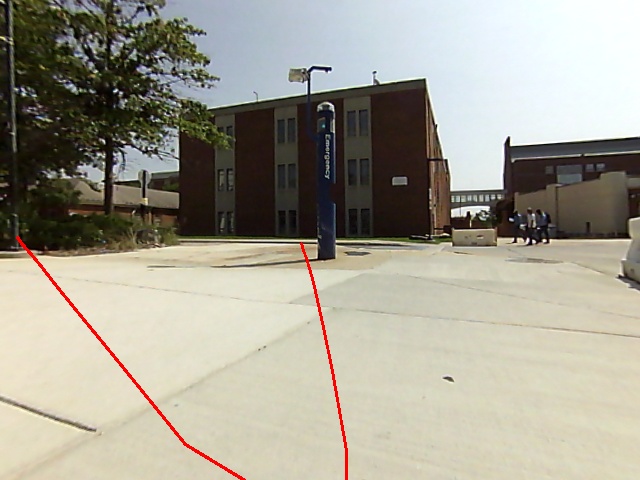} 
    &
    \includegraphics[height=.13\linewidth]{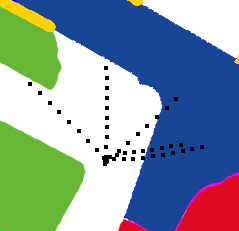}
    &
    \includegraphics[height=.13\linewidth]{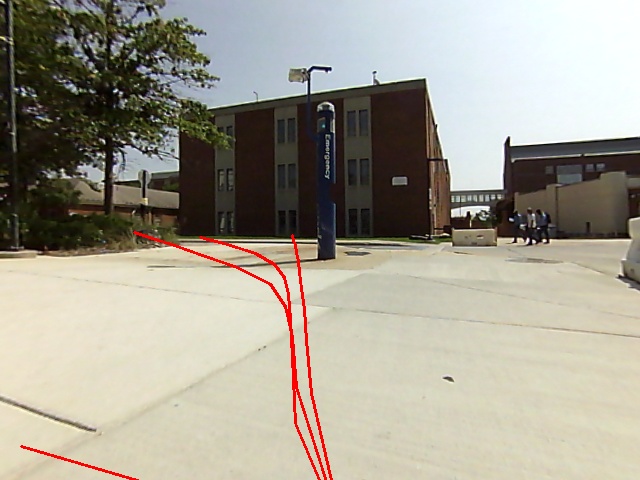} 
    &
    \includegraphics[height=.13\linewidth]{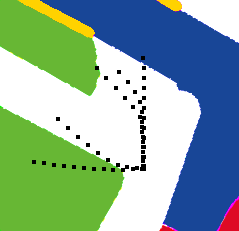} 
    \\ 

    \includegraphics[height=.13\linewidth]{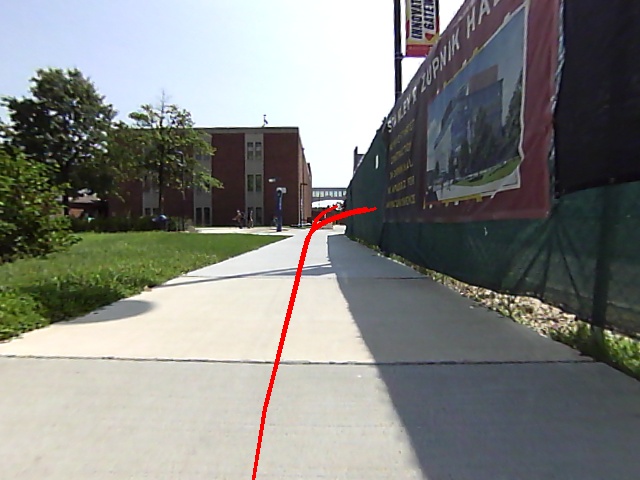}
    &
    \includegraphics[height=.13\linewidth]{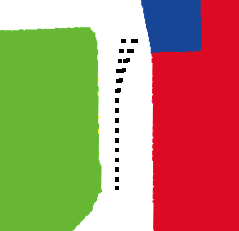} 
    &
    \includegraphics[height=.13\linewidth]{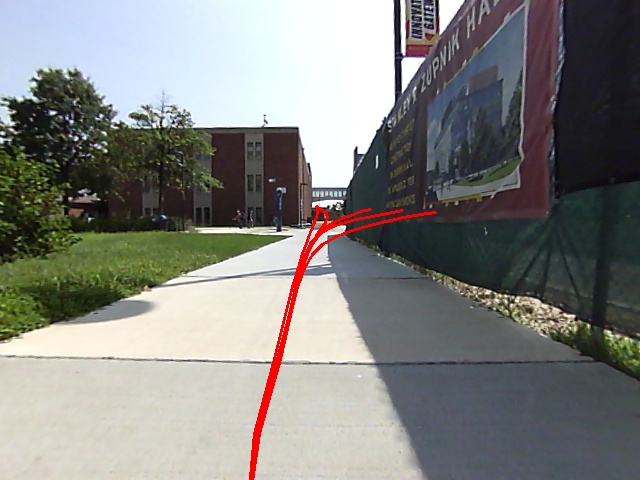} 
    &
    \includegraphics[height=.13\linewidth]{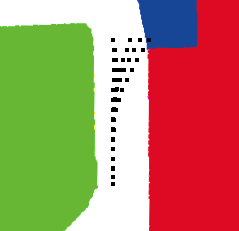} 
    &
    \includegraphics[height=.13\linewidth]{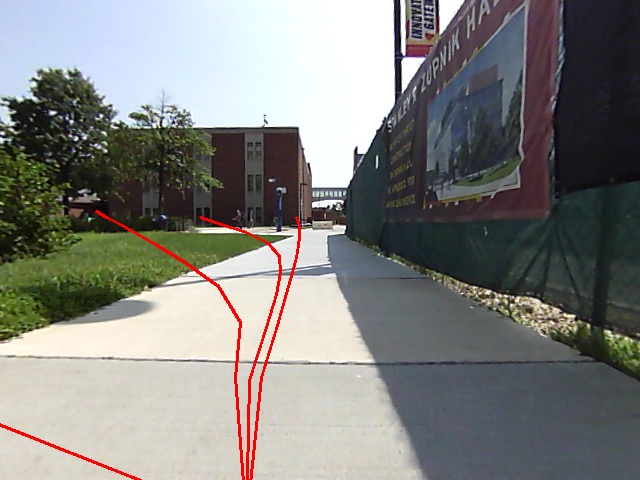} 
    &
    \includegraphics[height=.13\linewidth]{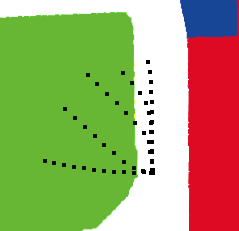} 
    \\ 
    
    \includegraphics[height=.13\linewidth]{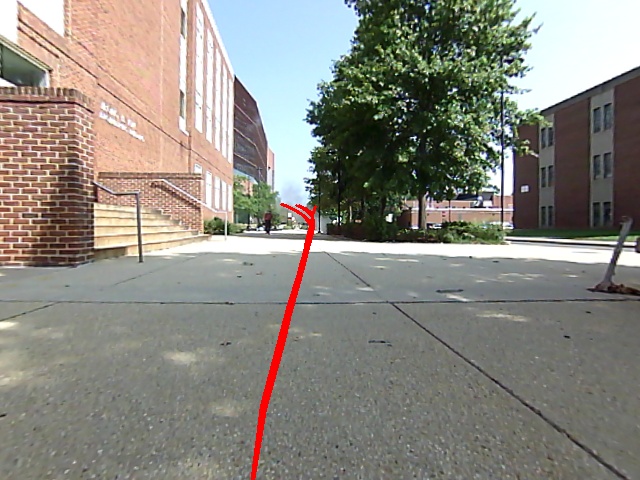} 
    &
    \includegraphics[height=.13\linewidth]{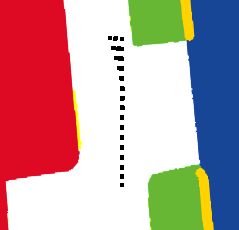} 
    &
    \includegraphics[height=.13\linewidth]{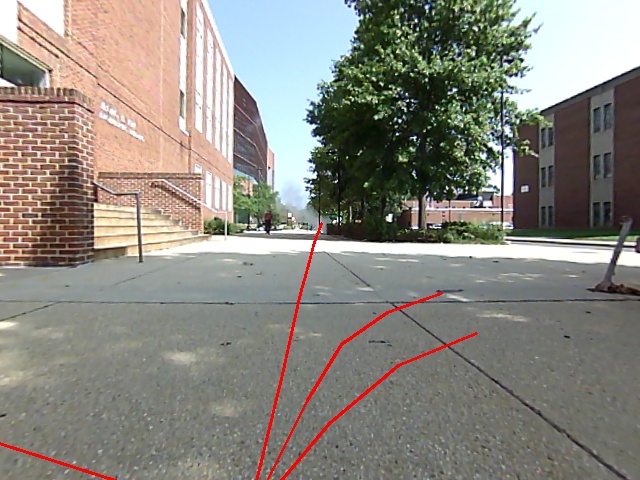} 
    &
    \includegraphics[height=.13\linewidth]{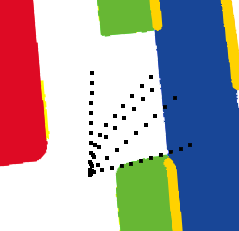} 
    &
    \includegraphics[height=.13\linewidth]{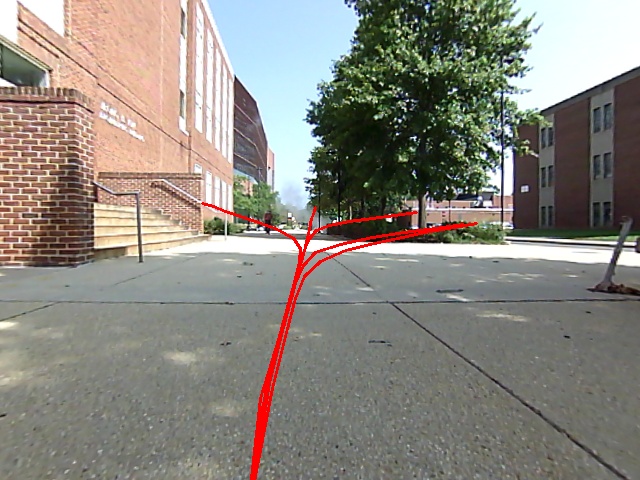} 
    &
    \includegraphics[height=.13\linewidth]{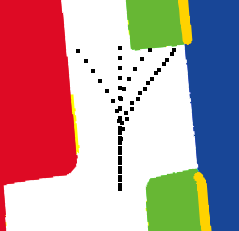} 
    \\ 
    
    
  \end{tabular}
  \caption{\textbf{T-MTG Generated Trajectories}: T-MTG generates trajectories to cover traversable areas across three levels: basic, agile, and legged traversability levels. Each level imposes different constraints and utilizes different types of robots for navigation. Basic traversability level contains only pedestrian roads for all types of robots, agile traversability level includes both pedestrian roads and vehicle roadways for fast-moving wheeled robots, and legged traversability level has pedestrian walkways and off-road terrains for legged robots. For each cell, the left side displays the generated trajectories (red solid lines) overlaid on the robot's RGB view image, while the right side shows the generated trajectories (black dotted lines) overlaid on the cropped multi-category traversability map. 
}
  \label{tb:extended_mtg}
  \vspace{-1.0em}
\end{figure*}

\endgroup